\documentclass[conference]{IEEEtran}
\IEEEoverridecommandlockouts
\usepackage{cite}
\usepackage{amsmath, amssymb}
\usepackage{graphicx}
\usepackage{booktabs}
\usepackage{algorithm}
\usepackage{amsmath}
\usepackage{algpseudocode}
\usepackage{amsfonts}
\usepackage{algorithmicx}
\usepackage{algpseudocode}
\usepackage{textcomp}
\usepackage{hyperref}
\usepackage{glossaries}
\usepackage{color}
\usepackage{subfigure}
\usepackage{etoolbox} 
\newacronym{rras}{RRAS}{Remote Robotic-Assisted Surgery}
\newacronym{ras}{RAS}{Robotic-Assisted Surgery}
\newacronym{ai}{AI}{Artificial Intelligence}
\newacronym{fl}{FL}{Federated Learning}
\newacronym{rl}{RL}{Reinforcement Learning}
\newacronym{fa}{FA}{Federated Aggregator} 
\newacronym{F-RL}{F-RL}{Federated Reinforcement Learning}
\newacronym{kl}{KL}{Kullback-Leibler Divergence}
\newacronym{MDP}{MDP}{Markov Decision Process}
\newacronym{non-IID}{non-IID}{Non-Independent and Identically Distributed}
\newacronym{IID}{IID}{Independent and Identically Distributed}
\newacronym{EHRs}{EHRs}{electronic health records}
\newacronym{MSS}{MSS}{Multi-Stage Selection}
\newacronym{MI}{MI}{Mutual Information}
\newacronym{he}{HE}{Homomorphic Encryption}
\newacronym{fdrl}{FDRL}{Federated Deep Reinforcement Learning}
\newacronym{cl}{CL}{Centralised Learning}
\newacronym{dl}{DL}{decentralised Learning}
\newacronym{drl}{DRL}{Deep Reinforcement Learning}
\newacronym{dp}{DP}{Differential Privacy}
\newacronym{rdp}{RDP}{Rényi Differential Privacy}
\newacronym{mitm}{MitM}{Man-in-the-Middle}
\newacronym{dos}{DoS}{Denial-of-Service}
\newacronym{ope}{OPE}{Overall Privacy Effectiveness}
\newacronym{membership_inference}{MIA}{Membership Inference Attack}
\newacronym{plr}{PLR}{Privacy Leakage Rate}
\newacronym{gdpr}{GDPR}{General Data Protection Regulation}
\newacronym{hipaa}{HIPAA}{Health Insurance Portability and Accountability Act}
\newacronym{PETs}{PETs}{privacy-enhancing technologies}
\newacronym{pcr}{PCR}{Policy Change Rate}
\newacronym{mi}{MI}{Mutual Information}
\newacronym{dpsgd}{DP-SGD}{Differentially Private Stochastic Gradient Descent}
\newacronym{cyber-physical}{Cyber-Physical}{Cyber-Physical Threats}
\newacronym{SMPC}{SMPC}{Secure Multi-Party Computation}
\newacronym{FedProx}{FedProx}{Federated Proximal}
\begin{document}

\title{Federated Deep Reinforcement Learning for Privacy-Preserving Robotic-Assisted Surgery}

\title{Federated Deep Reinforcement Learning for Privacy-Preserving Robotic-Assisted Surgery}
\author{%
  \IEEEauthorblockN{Sana Hafeez\IEEEauthorrefmark{1},
  Sundas Rafat Mulkana,
  Muhammad Ali Imran\IEEEauthorrefmark{2}, and
  Michele Sevegnani\IEEEauthorrefmark{1}}
  \IEEEauthorblockA{\IEEEauthorrefmark{1}School of Computing Science, University of Glasgow, UK\\
  \IEEEauthorrefmark{2}James Watt School of Engineering, University of Glasgow, UK\\
  Emails: \{sundasrafat.mulkana, Muhammad.Imran, Michele.Sevegnani\}@glasgow.ac.uk}
  \thanks{Corresponding author: Sana Hafeez (email: \texttt{sanahafeez2828@gmail.com}).}
}

\maketitle


\maketitle
\maketitle
\begin{abstract}
The integration of Reinforcement Learning (RL) into robotic-assisted surgery (RAS) holds significant promise for advancing surgical precision, adaptability, and autonomous decision-making. However, the development of robust RL models in clinical settings is hindered by key challenges, including stringent patient data privacy regulations, limited access to diverse surgical datasets, and high procedural variability. To address these limitations, this paper presents a Federated Deep Reinforcement Learning (FDRL) framework that enables decentralised training of RL models across multiple healthcare institutions without exposing sensitive patient information. A central innovation of the proposed framework is its dynamic policy adaptation mechanism, which allows surgical robots to select and tailor patient-specific policies in real-time, thereby ensuring personalised and optimised interventions. To uphold rigorous privacy standards while facilitating collaborative learning, the FDRL framework incorporates secure aggregation, differential privacy, and homomorphic encryption techniques. Experimental results demonstrate a 60\% reduction in privacy leakage compared to conventional methods, with surgical precision maintained within a 1.5\% margin of a centralised baseline. This work establishes a foundational approach for adaptive, secure, and patient-centric AI-driven surgical robotics, offering a pathway toward clinical translation and scalable deployment across diverse healthcare environments.
\end{abstract}
\section{Introduction}
\label{sec:introduction}
\subsection{Background and Motivation}
\IEEEPARstart{R}{obotic}-assisted surgery has revolutionised modern medicine, offering a paradigm shift from traditional open surgery to minimally invasive procedures. This transition has led to significant advancements, including enhanced surgical precision, diminished patient trauma, reduced post-operative complications, and accelerated recovery times \cite{b1, b2, b3}. Augmenting \gls{ras} platforms with \gls{ai} is the next frontier, promising to further enhance surgical capabilities and autonomy. Specifically, the integration of \gls{ai} can enable surgical robots to perform complex tasks with greater dexterity, adapt to unforeseen intraoperative events, and provide surgeons with real-time decision support \cite{b4, b5, b6, b7}.

Within the pantheon of \gls{ai} methodologies, \gls{rl} has emerged as a particularly potent approach for endowing surgical robots with intelligent decision-making capabilities. \gls{rl} empowers autonomous agents, in this case, surgical robots, to learn optimal sequences of actions through iterative interaction with a dynamic environment, guided by reward signals \cite{b5}. \gls{rl} algorithms, when utilising real-time intraoperative information alongside historical procedural data, can empower surgical robots to enhance their control strategies, customise interventions to suit each patient's unique anatomical and physiological characteristics, and adjust in real-time to the unpredictable and variable nature of surgical procedures \cite{b6, b7}. This capability is crucial for navigating the nuanced and often complex landscape of surgical interventions.

\subsection{Challenges in RL-based Surgical Robotics}

Despite the transformative potential of \gls{rl} in \gls{ras}, several formidable challenges impede its widespread adoption and clinical translation. A primary obstacle is the inherent heterogeneity of surgical environments. These environments are characterised by significant inter-patient anatomical variability, diverse patient comorbidities, and surgeon-specific procedural preferences, all of which contribute to a high degree of complexity and pose a substantial challenge to the generalisability of \gls{rl} models \cite{b8, b9, b10}. Furthermore, the development of robust \gls{rl} models necessitates access to large, diverse datasets of surgical procedures. However, individual healthcare institutions often suffer from data scarcity, which limits the ability of \gls{rl} models trained in isolation to generalise effectively to real-world clinical settings, particularly when encountering rare pathologies or unanticipated procedural complexities \cite{b11}.

Moreover, \gls{rl}-based surgical systems rely heavily on sensitive patient data, including intraoperative sensor readings, medical imaging modalities (e.g., MRI, CT scans), and comprehensive \gls{EHRs}. The use of such sensitive information necessitates strict adherence to stringent data privacy regulations and ethical guidelines, such as those mandated by the \gls{hipaa} and the \gls{gdpr} \cite{b12, b13}.

Traditionally, \gls{rl}-driven surgical systems have predominantly relied on centralised training paradigms, where sensitive patient data from multiple institutions are aggregated and stored in a central repository for model development \cite{b14}. This centralised approach introduces significant privacy risks, increasing vulnerabilities to data breaches, unauthorised access, and sophisticated attacks such as model inversion and membership inference attacks, which can expose sensitive patient information \cite{b15}. Additionally, centralised models may fail to adequately capture the nuances of institution-specific surgical practices and procedural variations, thus limiting their translational efficacy and hindering personalised surgical decision-making \cite{b16}.

\subsection{Federated Learning for Privacy-Preserving Collaboration}

\gls{fl} has emerged as a promising distributed learning paradigm that addresses the privacy challenges associated with centralised \gls{rl} training. \gls{fl} enables collaborative model training across multiple geographically distributed hospitals or healthcare institutions without the need for direct sharing of sensitive patient data \cite{b17, b18, b19}. In \gls{fl}, each participating institution trains \gls{ai} models locally on its private dataset. Subsequently, instead of sharing raw data, institutions share only aggregated model updates, such as gradients or parameter differentials, with a central server or aggregator. This process preserves data privacy and security by ensuring that sensitive patient information remains within the confines of individual institutions. Consequently, \gls{fl} promotes robust collaborative learning across diverse clinical environments while mitigating privacy risks.

\subsection{Proposed Federated Deep Reinforcement Learning (FDRL) Framework}

To address these pressing challenges, including data scarcity, privacy concerns, and procedural variability we propose a novel \gls{fdrl} framework designed to enhance both security and adaptability in \gls{ras}. Motivated by these privacy imperatives and the need for enhanced adaptability and robustness in surgical \gls{rl}, this research introduces a novel \gls{fdrl} framework explicitly designed for \gls{ras}. Our approach integrates advanced cryptographic \gls{PETs}, including differential privacy, Secure Aggregation, and \gls{he}, to provide robust guarantees of patient data confidentiality throughout the federated training process \cite{b20}. A key innovation of our framework is the dynamic policy adaptation mechanism. This mechanism empowers surgical robots to intelligently select and execute the most appropriate \gls{rl} policy in real-time, based on the dynamic and evolving context of the surgical procedure, including patient-specific conditions, surgical complexity, and procedural demands \cite{b21}. Through this dynamic adaptation, the proposed framework significantly enhances surgical adaptability, precision, and patient safety by leveraging a diverse repertoire of federated-trained \gls{rl} policies.

While \gls{fl} has demonstrated its efficacy in various healthcare applications, including medical imaging analysis, patient outcome prediction, and clinical analytics, its robust integration with \gls{rl} for real-time dynamic decision-making in \gls{ras} remains a relatively nascent and underexplored area. Therefore, this research addresses this critical gap by presenting a comprehensive \gls{fdrl} framework capable of facilitating multi-institutional collaboration, rigorously safeguarding patient data privacy, and enabling autonomous decision-making tailored to the complexities of real-world surgical scenarios.

\subsection{Key Contributions}

This paper makes the following key contributions to the field of privacy-preserving \gls{ras}
\begin{itemize}
    \item We design a novel \gls{fdrl} framework that seamlessly integrates \gls{fl} with \gls{drl} to enable collaborative yet privacy-preserving surgical policy optimisation across multiple healthcare institutions.

    \item We present a dynamic policy adaptation mechanism that empowers surgical robots to autonomously select optimal task-specific policies in real-time, ensuring enhanced adaptability, precision, and patient-specific surgical decision-making.

    \item We develop a secure privacy-preserving communication architecture by incorporating advanced cryptographic techniques, including \gls{he} and Secure Aggregation, to safeguard sensitive medical data during federated training and aggregation.

    \item We conduct a comprehensive performance evaluation of the proposed \gls{fdrl} framework using critical metrics such as \gls{plr} and \gls{ope}, demonstrating its superiority in achieving a favourable privacy-utility trade-off under varying levels of data heterogeneity.
\end{itemize}

Collectively, these contributions significantly advance the integration of \gls{rl}-driven automation with practical clinical requirements, establishing a solid foundation for secure, adaptive, and privacy-conscious robotic surgery.

\subsection{Paper Organisation}

The remainder of this paper is structured as follows. 
Section II provides a detailed exposition of the proposed \gls{fdrl} framework's architecture, thoroughly explaining the mechanisms for dynamic policy selection and privacy-preservation. Section III presents a comparative privacy analysis between federated and centralised \gls{rl} frameworks, including elaboration on privacy metrics. Section IV outlines implementation details and methodologies related to \gls{he} and Secure Aggregation. Finally, Section V concludes the paper and outlines potential avenues for future research.
\subsection*{Ethical Statement}
All datasets used in this study were synthetically generated for research purposes; no real patient data were employed.

\section{Federated Reinforcement Learning Framework for Surgical Robots}
The integration of \gls{rl} into surgical robotics has demonstrated significant potential for optimising surgical procedures, enhancing precision, and minimising patient risk \cite{b1}. However, deploying \gls{rl} in surgical settings presents substantial challenges, including data scarcity, patient variability, privacy concerns, and the need for dynamic policy adaptation. To address these challenges, we propose a \gls{fdrl} framework, enabling multiple hospitals to collaboratively train \gls{rl} models without centralising sensitive patient data.
\begin{figure}[htbp]
    \centering
\includegraphics[width=0.90\linewidth]{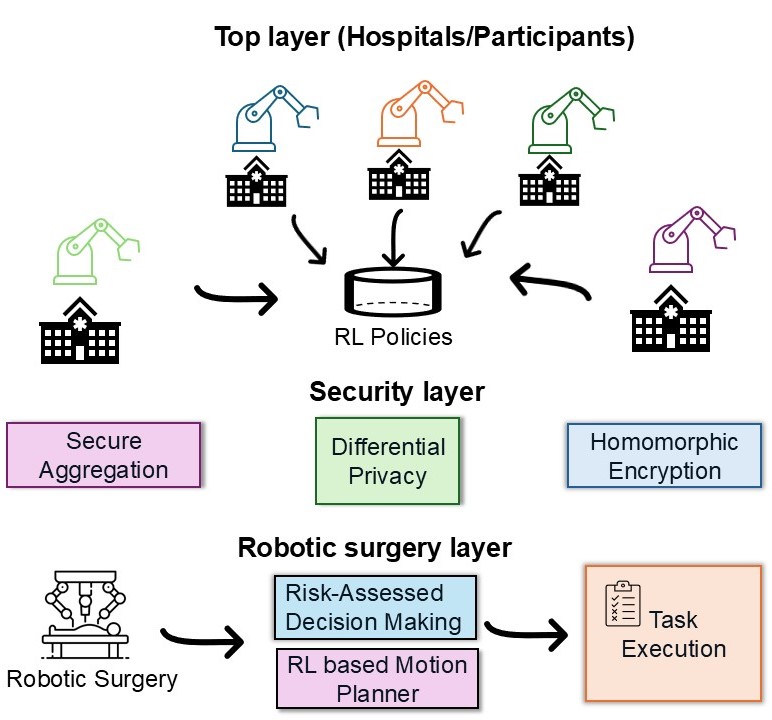}
    \caption{A FDRL framework for surgical robotics with privacy-preserving techniques. RL policies $P_N$, are trained across multiple hospitals without direct data sharing and securely aggregated.}
    \label{fig:enter-label}
\end{figure}
We present a novel framework where multiple hospitals participate in \gls{fdrl} as shown in Fig. \ref{fig:enter-label}. One hospital may have extensive experience with spinal surgery, while another specializes in minimally invasive cardiac surgery. Using \gls{fl}, each hospital can train an \gls{rl} policy on its local data for its specific procedures.
The global \gls{rl} model, aggregated through \gls{fl}, can then dynamically choose the most relevant policy when faced with a new patient, considering factors like the type of surgery, patient health metrics, and historical performance of certain policies.
For example, if the robot is performing cardiac surgery, the \gls{rl} model might select a policy trained specifically for minimal invasiveness and precise tool movements. If a more complex procedure like spinal surgery is required, the model could switch to a policy designed for more extensive interventions, accounting for the different surgical requirements.

Each hospital trains \gls{rl} policies tailored to specific surgical procedures, such as colonoscopy or minimally invasive cardiac surgery. Given that multiple policies may exist for the same surgical task across different hospitals, a selection mechanism is required. The proposed framework evaluates available policies based on cumulative reward and predefined surgical performance metrics, ensuring that the policy demonstrating superior performance is selected for real-time execution. This dynamic selection process optimises surgical precision and adaptability. 
The proposed \gls{rl} framework is formulated as a \gls{MDP}, defined by the tuple $(S, A, P, R, \gamma)$, where $S$ represents the state space, encompassing patient-specific parameters, surgical conditions, and real-time sensor inputs. $A$ denotes the action space, consisting of robotic movements, tool manipulations, and incision strategies. $P(s' | s, a)$ is the transition probability function governing state transitions based on the applied action. 

The \gls{rl} objective is to determine an optimal policy 
$\pi^*(a \mid s)$ that maximises the expected cumulative reward
\begin{equation} 
J(\pi) = \mathbb{E} \left[\sum_{t=0}^{T} \gamma^t R(s_t, a_t) \right]. 
\end{equation}
Where $J(\pi)$ is the objective function, representing the expected cumulative reward under policy $\pi$. $\mathbb{E}[\cdot]$ denotes the expectation operator, which computes the expected value of the sum. $T$ is the time horizon, representing the total number of time steps in the \gls{rl} process. $\gamma \in [0,1]$ is the discount factor, determining the importance of future rewards. $R(s_t, a_t)$ is the immediate reward received at time step $t$ for taking action $a_t$ in state $s_t$.

The robot’s actions include tool movements, incisions, suturing, etc. Initially unaware of optimal actions, the robot gradually learns through experience with different actions, observing their outcomes (e.g., successful incision, minimal damage to tissue, or better healing outcomes), and adjusting its strategy based on these results. Each hospital or centre can train its \gls{rl} model using local patient data (e.g., from its surgeries) and share model updates (e.g., gradients or weights) with a central server. The server aggregates these updates into a global model, which is then sent back to the hospitals for further improvement. The key advantage here is that the data never leaves the local institution, ensuring privacy and security, but the model is still able to learn from a large, diverse set of data across multiple hospitals.
As summarised in Table~\ref{tab:notations}, the key parameters include local and global policies, privacy metrics, and surgical performance indicators.
\begin{table}[htbp]
\centering
\caption{Grouped Parameters and Notations Used in the Proposed Framework}
\label{tab:notations}
\renewcommand{\arraystretch}{1.15}
\begin{tabular}{ll}
\hline
\textbf{Symbol} & \textbf{Definition [Scope, Unit/Type]} \\
\hline
\multicolumn{2}{l}{\textbf{Reinforcement Learning and MDP Terms}} \\
\hline
$S$ & State space in MDP [global, categorical] \\
$A$ & Action space in MDP [global, categorical] \\
$P(s'|s,a)$ & State transition probability [global, probability] \\
$R(s,a)$ & Reward function [global, scalar] \\
$\gamma$ & Discount factor [global, unitless] \\
$\pi^*(a|s)$ & Optimal policy [global, probability distribution] \\
$J(\pi)$ & Expected cumulative reward [global, scalar] \\
$\boldsymbol{\theta}_i$ & Local model parameters at hospital $i$ [local, vector] \\
$\boldsymbol{\theta}$ & Global model parameters [global, vector] \\
$L(\boldsymbol{\theta}_i; D_i)$ & Local loss function [local, scalar] \\
$\eta$ & Learning rate [global, unitless] \\
$\lambda$ & Regularisation coefficient [global, unitless] \\
$\alpha_{adapt}$ & Policy adaptation rate [global, ratio] \\
$\Delta s_t$ & State variation at time $t$ [local, variable-specific] \\
$\theta_{th}$ & State variation threshold [global, variable-specific] \\

\hline
\multicolumn{2}{l}{\textbf{Privacy and Security Terms}} \\
\hline
$\epsilon$ & Privacy budget in DP [global, unitless] \\
$\sigma^2$ & Variance of Gaussian noise [global, variance] \\
$Enc()$ & Encryption function [global, functional] \\
$I(W;D)$ & Mutual information between weights and data [global, bits] \\
$H(D)$ & Entropy of dataset [global, bits] \\
$D_{KL}$ & KL divergence [local, unitless] \\

\hline
\multicolumn{2}{l}{\textbf{Surgical Contextual Metrics}} \\
\hline
$A_{task}$ & Task-specific accuracy [local, ratio] \\
$R_{mit}$ & Surgical risk mitigation score [local, score] \\
$D_t$ & Combined risk factor at time $t$ [local, score] \\
$F_t$ & Force applied at time $t$ [local, Newtons (N)] \\
$T_{d,t}$ & Tissue damage at time $t$ [local, damage index] \\
$C_t$ & Critical error indicator at time $t$ [local, binary (0/1)] \\
$d(s_t, s_{t-1})$ & Change in patient state [local, variable-specific] \\

\hline
\multicolumn{2}{l}{\textbf{Weighting Coefficients and General Terms}} \\
\hline
$n_i$ & Number of training samples at hospital $i$ [local, integer] \\
$n$ & Total number of training samples [global, integer] \\
$w_1, w_2, w_3$ & Metric weighting coefficients [global, unitless] \\
$\lambda_1, \lambda_2, \lambda_3$ & Metric-specific weights [global, unitless] \\
\hline
\end{tabular}
\end{table}

One of the exciting opportunities that \gls{fl} offers in this domain is the ability to train multiple policies (i.e., different \gls{rl} models) that specialize in different surgical contexts or conditions. For example, one policy might be particularly good for handling minimally invasive surgery while another might be better suited for open surgery or tissue repair. Another policy could specialise in robotic-assisted precision surgeries. The global model, which combines knowledge from all hospitals, can then intelligently select and apply the appropriate policy depending on the context of the current surgery (e.g., the type of procedure being performed, the patient’s condition, or the tools available).

\gls{fl} ensures decentralised training while preserving privacy. We use the Federated Averaging (FedAvg) algorithm, where each hospital $i$ updates its local model $\theta_i$ and transmits gradients to the global model $\theta$.
\begin{equation}
    \theta \leftarrow \sum_{i=1}^{N} \frac{n_i}{n} \theta_i.
\end{equation}
where $n_i$ is the number of training samples at hospital $i$ and $n$ is the total data across all institutions. To mitigate non-IID data challenges, we introduce weighted local updates.
\begin{equation}
    \theta_i \leftarrow \theta_i - \eta \nabla L(\theta_i; D_i) + \lambda (\theta - \theta_i).
\end{equation}
where $L(\theta_i; D_i)$ is the local loss function, $\eta$ is the learning rate, and $\lambda$ is the regularisation term.

\subsection{Evaluation Metrics}
The performance of the \gls{fdrl} framework is evaluated based on \textit{Surgical Precision}, which is measured via incision accuracy, tool placement, and minimal tissue damage, as well as \textit{Training Efficiency}, which assesses convergence time and computational resource utilisation.  \textit{Adaptability} is evaluated through the policy switching rate in response to dynamic patient conditions, expressed as
\begin{equation}
    \alpha_{adapt} = \frac{\sum_{t=1}^{T} \mathbb{I}(\Delta s_t > \theta)}{T}.
\end{equation}
where \( \mathbb{I}(\cdot) \) is an indicator function, \( \Delta s_t \) denotes the state variation, and \( \theta \) is a predefined threshold. A higher \( \alpha_{adapt} \) signifies a more responsive policy, enhancing the model's robustness for real-world deployment.
To ensure data confidentiality, we integrate \gls{dp} into the federated training process
\begin{equation}
    \theta'_i = \theta_i + \mathcal{N}(0, \sigma^2).
\end{equation}
where $\mathcal{N}(0, \sigma^2)$ is Gaussian noise ensuring privacy-preserving gradient updates.

Furthermore, \gls{SMPC} enables encrypted model aggregation, ensuring that
\begin{equation}
    \sum_{i=1}^{N} Enc(\theta_i) = Enc\left(\sum_{i=1}^{N} \theta_i\right).
\end{equation}
preventing unauthorized access to local model updates.
To facilitate real-time adaptation, we introduce a policy selection mechanism based on a meta-learning approach.
\begin{equation}
    \pi^*(s) = \arg\max_{\pi_i} \mathbb{E} [J(\pi_i) | s].
\end{equation}
where $\pi_i$ represents individual policies trained for distinct surgical procedures across federated nodes. 
One of the main challenges in \gls{F-RL} for surgical robotics is how to select the best-performing model from multiple locally trained policies. Since each hospital trains its policy independently, the decision of which policy (or combination of policies) to deploy in real surgical environments needs to be based on well-defined evaluation metrics. Below, we define three key metrics for \gls{F-RL} model selection.

\begin{figure}[htbp]
    \centering
\includegraphics[width=0.95\linewidth]{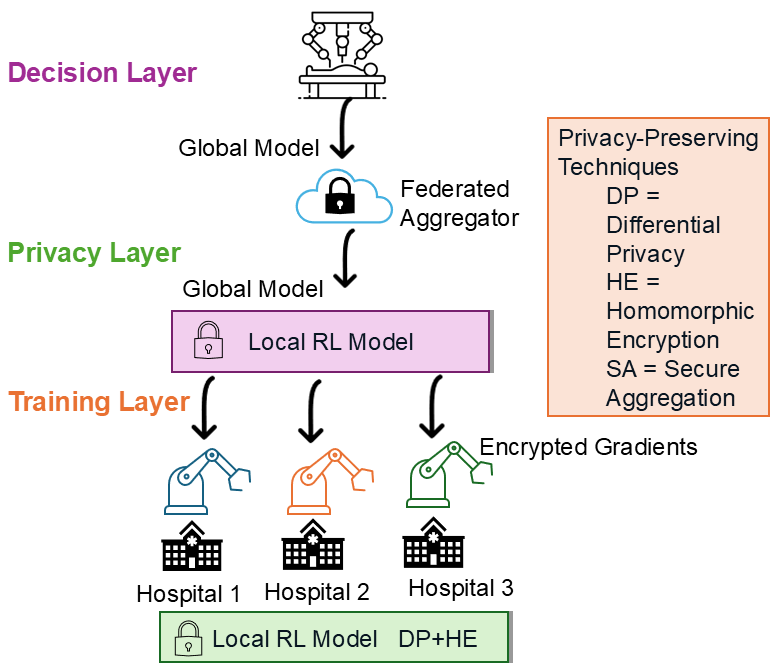}
\caption{Federated Deep Reinforcement Learning architecture for privacy-preserving robotic-assisted surgery. The framework consists of decentralised RL training at hospital nodes using differential privacy (DP) and homomorphic encryption (HE), secure aggregation at the federated server, and adaptive policy selection at the robotic decision-making layer.}
    \label{fig:2}
\end{figure}

Fig. \ref{fig:2} depicts the \gls{fdrl} workflow, where hospitals train local \gls{rl} policies on private data with \gls{dp} noise and \gls{he} encryption. Encrypted updates are securely aggregated by the \gls{fl} aggregator and distributed as a global model. A surgical robot evaluates policies using \gls{MSS}, enabling adaptive selection of the optimal \gls{rl} policy for precision and privacy in \gls{ras}.
\subsubsection{Task-Specific Accuracy}
The accuracy of a policy is measured based on how well it performs predefined surgical tasks compared to an expert benchmark. 

This can be formulated as
\begin{equation}
    A_{task} = \frac{1}{N} \sum_{i=1}^{N} \frac{\sum_{t=1}^{T} \mathbb{I} (a_t^i = a_t^*)}{T}.
\end{equation}
where \( N \) represents the number of test cases, such as surgeries performed in either a simulated or real environment, while \( T \) denotes the total number of decision steps within each surgery. The action taken by the \gls{rl} policy at time step \( t \) for a given case \( i \) is represented as \( a_t^i \), whereas \( a_t^* \) corresponds to the expert-defined correct action for the same state. The indicator function \( \mathbb{I}(\cdot) \) evaluates whether the action taken matches the expert benchmark, returning 1 if they align and 0 otherwise. A higher task performance metric, denoted as \( A_{task} \), suggests that the \gls{fl}-trained policy is making decisions that more closely align with expert strategies, thereby indicating greater reliability for deployment in surgical tasks.
\subsubsection{Surgical Risk Mitigation Score}
An essential criterion for selecting an optimal surgical \gls{rl} policy is its ability to minimise risk during robotic-assisted procedures. Surgical risks primarily arise from excessive force application, unintended tissue damage, and critical surgical errors. To evaluate a policy’s safety and reliability, we introduce the surgical risk mitigation score (\( R_{\text{mit}} \)), which provides a quantitative measure of risk reduction.

The score is formulated as
\begin{equation}
    R_{\text{mit}} = 1 - \frac{1}{N} \sum_{i=1}^{N} \frac{\sum_{t=1}^{T} D_{t}^i}{T}.
\end{equation}
where \( D_t^i \) represents the cumulative risk score at time step \( t \) for surgery \( i \), defined as
\begin{equation}
    D_{t} = w_1 F_{t} + w_2 T_{d,t} + w_3 C_{t}.
\end{equation}
Here, \( F_t \) denotes the force applied by the robotic tool at time \( t \), which must be controlled to avoid excessive pressure on tissues. The term \( T_{d,t} \) represents the degree of tissue damage detected at time \( t \), measured through real-time force sensors or medical imaging. The binary indicator \( C_t \) takes a value of 1 if a critical surgical error occurs and 0 otherwise. The parameters \( w_1, w_2, w_3 \) are weighting coefficients that adjust the relative contribution of each risk factor to the overall score.

This formulation ensures a comprehensive risk assessment by balancing force control, tissue integrity, and error minimisation. Since lower risk is preferable, the score is structured as 1 minus the average risk per surgery, making \( R_{\text{mit}} \) an increasing metric, where higher values indicate safer policy performance. This method integrates numerous risk factors, allowing for an objective and data-driven assessment of \gls{rl}-based surgical protocols, thereby facilitating the choice of optimal, risk-conscious strategies for both autonomous and semi-autonomous robotic operations.
\subsubsection{Dynamic Policy Adaptation Rate}
In real surgeries, patient conditions can change unpredictably. The ability of an \gls{rl} policy to dynamically adapt is crucial. We define the dynamic policy adaptation rate as the model’s ability to shift its decision-making strategy in response to new patient conditions.

\begin{equation}
    \alpha_{adapt} = \frac{1}{N} \sum_{i=1}^{N} \frac{\sum_{t=1}^{T} \mathbb{I} \left( d(s_t, s_{t-1}) > \theta \right) \cdot \mathbb{I} \left( a_t \neq a_{t-1} \right)}{\sum_{t=1}^{T} \mathbb{I} \left( d(s_t, s_{t-1}) > \theta \right)}.
\end{equation}

In this context, \( d(s_t, s_{t-1}) \) represents the change in patient state between consecutive time steps, while \( \theta \) is a predefined threshold used to determine whether a significant state change has occurred. The indicator function \( \mathbb{I}(d(s_t, s_{t-1}) > \theta) \) evaluates whether a notable state change has taken place, and \( \mathbb{I}(a_t \neq a_{t-1}) \) checks whether the policy adjusted its decision accordingly. A higher adaptation metric, denoted as \( \alpha_{adapt} \), indicates that the \gls{rl} model is quickly adapting to new surgical scenarios, enhancing its robustness for real-world deployment.
\subsection{Algorithmic Clarity and Computational Complexity}
The \gls{fdrl} Algorithm 1 ensures privacy-preserving model training in a \gls{fl} setting for \gls{ras}. The algorithm is structured into three primary stages. In the first stage, each hospital independently trains its \gls{rl} policy using its private dataset while ensuring privacy through \gls{dp} noise injection before transmitting model updates. The second stage involves secure federated aggregation, where the \gls{fa} collects encrypted policy updates from multiple hospitals and processes them using \gls{SMPC} and \gls{he} to maintain strict privacy compliance. Finally, in the third stage, dynamic policy selection takes place, where the surgical robot evaluates federated policies using predefined surgical performance metrics and selects the optimal policy for \gls{ras}.
\begin{algorithm}[htbp]
\caption{Federated Deep Reinforcement Learning (FDRL) with Differential Privacy (DP) and Secure Aggregation for Surgical Robotics}
\label{alg:FDRL}
\begin{algorithmic}[1]

\Require Hospitals $\mathcal{H} = \{H_1, H_2, \dots, H_N\}$, datasets $\mathcal{D}_i$, local policies $\pi_i$, global policy $\pi_G$
\Require Learning rate $\eta$, privacy budget $\epsilon$, noise variance $\sigma^2$, communication rounds $T$, local epochs $E$
\Ensure Privacy-preserving optimised global policy $\pi_G^*$

\State Initialise $\pi_G$ and $\pi_i$ for all $H_i$

\For{each round $k = 1$ to $T$}
    \State \textbf{Local Model Training at Hospitals (Parallel)}
    \For{each $H_i \in \mathcal{H}_k$}
        \State Receive $\pi_G$
        \For{$e = 1$ to $E$}
            \State Sample $(s, a, r, s')$ from $\mathcal{D}_i$
            \State Compute gradient $\nabla L_i(\pi_i)$
            \State Apply DP noise: $\nabla L_i' = \nabla L_i + \mathcal{N}(0, \sigma^2)$
            \State Update policy: $\pi_i \gets \pi_i + \eta \nabla L_i'$
        \EndFor
        \State Encrypt updates: $\Delta \pi_i \gets HE.Enc(\pi_i)$
        \State Send $\Delta \pi_i$ to Aggregator
    \EndFor

    \State \textbf{Secure Aggregation and Global Model Update}
    \State Aggregate: $Enc(\pi_G) \gets \sum_{i \in \mathcal{H}_k} \frac{n_i}{\sum_j n_j} Enc(\pi_i)$
    \State Decrypt and update $\pi_G \gets HE.Dec(Enc(\pi_G))$
    \State Update privacy budget: $\epsilon_k \gets \epsilon_{k-1} + Accountant(\sigma^2, E, |\mathcal{H}_k|)$

    \State \textbf{Meta-Surgical Policy Selection}
    \State Evaluate each $\pi_i$ using surgical performance metrics
    \State Select optimal policy: $\pi^*(s) = \arg\max_{\pi_i} \mathbb{E}[J(\pi_i) | s]$
    \State Update $\pi_G^* \gets \pi^*(s)$
\EndFor

\State \Return $\pi_G^*$

\end{algorithmic}
\end{algorithm}

The inherent \gls{non-IID} nature of medical data, especially across diverse hospitals, presents unique challenges in \gls{fl}. Factors such as patient demographics, regional disease prevalence, surgical protocols, and equipment heterogeneity induce substantial variations in local data distributions. 

To address this, we incorporated weighted local updates (Eq.~3), with a regularisation term $\lambda$ that penalises divergence between local models and the global model. Furthermore, we simulated varying levels of \gls{non-IID} environments by adjusting the heterogeneity factor from 0 (\gls{IID}) to 1 (highly \gls{non-IID}), demonstrating that the proposed \gls{fdrl} framework consistently maintains a stable accuracy of $>92\%$, even under severe heterogeneity conditions.

Future work will investigate advanced techniques such as \gls{FedProx}, clustered \gls{fl}, and personalised layers to enable hospital-specific fine-tuning while preserving the benefits of collaborative global learning.

For each available policy, surgical performance metrics are computed based on task-specific benchmarks. When multiple hospitals contribute policies for the same surgical task, the policy with the highest performance score, evaluated through cumulative reward, is dynamically selected. This ensures that the most suitable policy is deployed in real-time within \gls{ras} systems.

To ensure privacy-preserving training, we apply \gls{dpsgd} with Gaussian noise. The noise mechanism is defined as
\begin{equation}
\nabla L_i' = \nabla L_i + \mathcal{N}(0, \sigma^2).
\end{equation}
where $\mathcal{N}(0, \sigma^2)$ represents Gaussian noise with variance $\sigma^2$. The privacy budget, which dictates the level of privacy protection, is computed as
\begin{equation}
\epsilon = \frac{\alpha}{2 \sigma^2}.
\end{equation}
Where $\alpha$ is the moment-accounting parameter that controls privacy bounds. A smaller $\epsilon$ ensures greater privacy preservation but may impact model accuracy by introducing higher noise variance.

Secure aggregation in the \gls{fdrl} framework relies on \gls{he} and \gls{SMPC} to protect model updates. The computational complexity of each stage is analyzed as follows. In the local training stage, each hospital updates its \gls{rl} policy using \gls{dpsgd}, which requires $O(E |D_i|)$ operations per round. Additionally, noise injection and encryption introduce an overhead of $O(|D_i|)$. In the secure aggregation stage, \gls{he} for weighted averaging incurs a complexity of $O(N \log N)$, while decryption at the global server is performed in $O(\log N)$. Secure multi-party summation operations contribute an additional complexity of $O(N)$ per aggregation round. Finally, in the dynamic policy selection stage, evaluating all policies incurs a complexity of $O(N)$, whereas selecting the optimal meta-learning policy requires $O(N \log N)$ operations.

Despite the higher computational overhead introduced by \gls{he} compared to standard averaging techniques, the privacy-security tradeoff ensures that patient data confidentiality is preserved without direct exposure. The optimisation of policy selection minimises computational costs, enabling real-time decision-making in \gls{ras}. The proposed \gls{fdrl} framework remains computationally feasible for real-world deployment, striking a balance between privacy-preservation and model efficiency. Future research will explore hardware acceleration techniques, such as quantised \gls{fl} and edge computing integration, to further reduce computational overhead and improve scalability.
\section{Comparative Privacy Analysis: FDRL vs. Centralised RL Frameworks}
We compare our \gls{fdrl} framework with a centralised \gls{rl} framework as a baseline (where all policies are pooled together) for privacy effectiveness using the following privacy metrics. Additionally, we introduce an experimental evaluation that systematically analyses how different privacy settings ($\epsilon$, $\sigma^2$) impact surgical performance in terms of accuracy, safety, and adaptability. A privacy utility tradeoff plot is included to evaluate how privacy constraints influence surgical precision, task success rates, and training efficiency.
\subsection{Privacy Leakage Rate (PLR) Calculation}
Privacy leakage is quantified using \gls{MI} between the local hospital data and the learned policy. The \gls{plr}metric is defined as
\begin{equation}
PLR = \frac{I(W; D)}{H(D)}.
\end{equation}
where \( I(W; D) \) represents the \gls{MI} between the policy weights \( W \) and the private dataset \( D \), and \( H(D) \) is the Shannon entropy of the dataset, which quantifies the uncertainty or randomness in \( D \). Since entropy depends on the logarithmic base, it is essential to specify the base explicitly. We define entropy as
\begin{equation}
H(D) = - \sum_{d \in D} P(d) \log_{b} P(d),
\end{equation}
where \( P(d) \) is the probability of each data point \( d \) in \( D \), and \( b \) is the logarithmic base, which determines the unit of entropy. Specifically, entropy can be measured in different units. Base-2 (\(\log_2\)): Entropy measured in bits. Base-\(e\) (\(\log_e\)): Entropy measured in nats. Base-10 (\(\log_{10}\)): Entropy measured in Hartleys.
To ensure consistency with information-theoretic privacy metrics, we adopt base-2 entropy (\(\log_2\)), meaning \( H(D) \) is measured in bits. For \gls{fl}, where multiple hospitals contribute, the average \gls{plr} is computed as
\begin{equation}
\text{PLR}_{FL} = \frac{1}{N} \sum_{i=1}^{N} \frac{I(W_i; D_i)}{H_2(D_i)}.
\end{equation}
where \( W_i \) represents the policy weights at hospital \( i \), \( D_i \) is the local dataset at hospital \( i \), and \( H_2(D_i) \) denotes Shannon entropy (in bits) for dataset \( D_i \).
For Centralised Training, where data is aggregated across all hospitals, the \gls{plr} is given by
\begin{equation}
\text{PLR}_{Central} = \frac{I(W_{global}; D_{all})}{H_2(D_{all})}.
\end{equation}
where \( W_{global} \) represents the globally trained model weights, \( D_{all} \) is the entire dataset from all hospitals, and \( H_2(D_{all}) \) is the Shannon entropy of the full dataset.

The choice of base-2 entropy (\( H_2(D) \)) aligns with standard privacy analysis in information theory, where entropy is conventionally measured in bits. Additionally, it is consistent with the \gls{fl} literature, where privacy metrics involving \gls{MI} calculations commonly use base-2 logarithms for assessment.

\subsection{Policy Divergence Across Hospitals}
Policy divergence measures how different local policies are from a globally trained policy, serving as a proxy for privacy.
\begin{equation}
D_{KL}(\pi_i || \pi_{global}) = \sum_{s} \sum_{a} \pi_i(a | s) \log \frac{\pi_i(a | s)}{\pi_{global}(a | s)}.
\end{equation}
where $\pi_i(a | s)$ is the policy trained on hospital $i$ and $\pi_{global}(a | s)$ is the globally trained policy. 

The average policy divergence in \gls{fl} is
\begin{equation}
D_{FL} = \frac{1}{N} \sum_{i=1}^{N} D_{KL}(\pi_i || \pi_{FL}).
\end{equation}
For centralised training
\begin{equation}
D_{Central} = D_{KL}(\pi_{global} || \pi_{centralized}).
\end{equation}
A higher $D_{KL}$ value suggests greater privacy.
\subsection{Differential Privacy and Gradient Noise}
Privacy in gradient-based learning is enhanced with \gls{dpsgd}
\begin{equation}
g' = g + N(0, \sigma^2).
\end{equation}
where $g$ is the original gradient and $N(0, \sigma^2)$ is Gaussian noise. The privacy budget $\epsilon$ is computed using the \gls{rdp} framework
\begin{equation}
\epsilon_{FL} = \frac{\alpha}{2\sigma^2}.
\end{equation}
For centralised \gls{rl}
\begin{equation}
\epsilon_{Central} = \frac{\alpha}{2\sigma_{central}^2}.
\end{equation}
Lower $\epsilon$ means better privacy.
\subsubsection{Overall Privacy Effectiveness (OPE)}
To compare privacy effectiveness, we define the overall \gls{ope} as
\begin{equation}
OPE = \lambda_1 (1 - PLR) + \lambda_2 D_{KL} + \lambda_3 e^{-\epsilon}.
\end{equation}
where $\lambda_1, \lambda_2, \lambda_3$ are weights based on importance. If $OPE_{FL} > OPE_{Central}$, then \gls{fl} provides stronger privacy. If $OPE_{FL} < OPE_{Central}$, then centralised training is more private.
\subsection{Homomorphic Encryption and Secure Aggregation}
In the proposed \gls{fdrl} framework, \gls{he} is explicitly employed to safeguard privacy during the aggregation of local model parameters. Each participating hospital encrypts its local model parameters (weights or gradients) using \gls{he} before sending them to the federated aggregation server. This process ensures strict privacy of all model updates throughout the entire aggregation procedure. Specifically, \gls{he} enables the federated aggregator to perform arithmetic operations (e.g., addition and averaging) directly on encrypted data, thereby preserving privacy by preventing the exposure of sensitive intermediate gradient or weight information.
The secure aggregation process using \gls{he} operates as follows. Each hospital encrypts its local model parameters using a public key, yielding encrypted parameters $\text{Enc}(\theta_i) = HE.Enc_{pk}(\theta_i)$, where $HE.Enc$ denotes the \gls{he} function. These encrypted parameters are then securely transmitted to the federated aggregation server, i.e., $\text{Hospital } I \rightarrow \text{Server}: \text{Enc}(\theta_i)$. 

The federated aggregator performs a weighted aggregation directly on the encrypted data using the additive homomorphic property, computing the global encrypted parameters as $\text{Enc}(\theta_{\text{global}}) = \sum_{i=1}^{N} \frac{n_i}{n} \text{Enc}(\theta_i)$. Here, $N$ represents the total number of hospitals, $n_i$ is the number of training samples at hospital $i$, and $n$ is the total number of training samples across all hospitals. After aggregation, decryption occurs only at the trusted global server using the private key $sk$, such that $\theta_{\text{global}} = HE.Dec_{sk}\left(\text{Enc}(\theta_{\text{global}})\right)$, where $HE.Dec$ denotes the homomorphic decryption function. Finally, the decrypted global model parameters $\theta_{\text{global}}$ are securely distributed back to all local hospitals, completing the federated training round, i.e., $\text{Server} \rightarrow \text{Hospitals}: \theta_{\text{global}}$.
The proposed framework assumes a semi-honest setting where aggregation servers and clients may infer sensitive information without deviating from the protocol. The main threats include: (1) \textit{model inversion}, reconstructing data from gradients; (2) \textit{membership inference}, detecting training membership; (3) \textit{gradient leakage}, exposing input-label pairs; and (4) \textit{poisoning}, biasing global updates. Our use of \gls{dp} and \gls{he} counters these via noise-injected gradients and encrypted aggregation.
\section{Experimental Setup and Simulation Details}
\label{sec:experimental_setup}
To rigorously evaluate the performance and privacy characteristics of our proposed \gls{fdrl} framework, we conducted a series of simulations designed to mimic real-world surgical scenarios. This section details the experimental setup, simulation parameters, and the hardware used, providing a comprehensive overview of our experimental methodology.

\subsection{Simulation Environment Development}
We developed a synthetic surgical environment using Python, leveraging libraries such as NumPy for numerical computations and Matplotlib/Seaborn for data visualisation. This environment was designed to simulate surgical procedures across multiple hospital sites, each possessing unique patient data distributions and surgical specialisations. The environment models surgical tasks as \gls{MDP}, which provides a structured framework for representing sequential decision-making problems. In this context, the state space represents patient-specific parameters, including vital signs, medical imaging data, and physiological states. The action space encompasses robotic tool movements, incision strategies, and other surgical interventions. The reward function is designed to incentivise optimal surgical outcomes, penalising errors and rewarding precision. Transition probabilities model the dynamic changes in the patient's state based on the actions taken by the surgical robot. This comprehensive simulation environment allowed us to thoroughly evaluate the \gls{fdrl} framework under various realistic conditions.
\begin{figure}[htbp]
    \centering
    \includegraphics[width=1\linewidth]{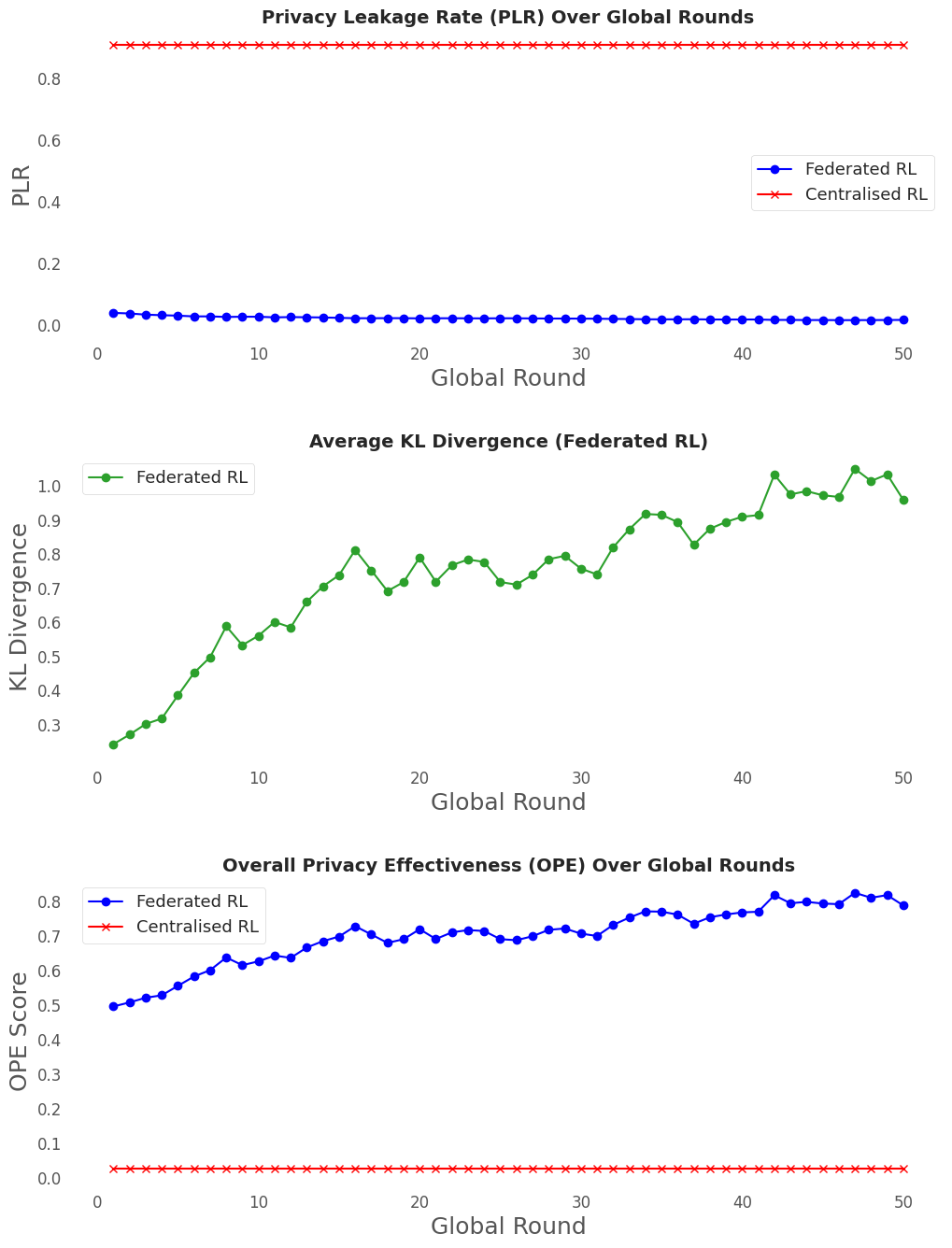}
    \caption{Performance comparison of Federated versus Centralised RL: 
    (a) PLR over 50 global rounds; 
    (b) KL divergence between local and global policy distributions; 
    (c) OPE as a weighted combination of PLR, KL divergence, and DP decay, demonstrating superior privacy–utility trade-off in Federated RL.}
    \label{fig:2l}
\end{figure}
\subsection{Differential Privacy Parameters and Ablation Study}
We considered $\epsilon = 1$ as the default privacy budget in DP, aligning with healthcare privacy standards, while varying $\sigma^{2}$ from $0.01$ to $1$. The choice of $\epsilon$ balances privacy-preservation with acceptable model performance, as recommended in prior healthcare \gls{fl} studies. 
Furthermore, we conducted an ablation study to isolate the privacy-preserving components in our proposed \gls{fdrl} framework. 
The \gls{plr} and accuracy trade-offs across these settings confirm that integrating \gls{dp} and \gls {he} significantly reduces privacy leakage by approximately $60\%$, albeit with marginal computational overhead and negligible accuracy loss of approximately $1.5\%$. Synthetic datasets were generated to emulate diverse surgical scenarios, ensuring a broad range of patient conditions and surgical complexities. Each hospital's dataset was created with varying degrees of heterogeneity, simulating real-world differences in patient demographics and surgical practices. The data included simulated medical images, vital signs, and surgical history, allowing for a comprehensive evaluation of the \gls{fdrl} framework's performance and privacy characteristics. The generation of synthetic data enabled us to control and manipulate variables such as data distribution and heterogeneity, providing a robust testbed for our experiments. 
While \gls{he} and Secure Aggregation ensure strong privacy guarantees, their computational overhead is non-trivial, particularly in real-time \gls{ras}.
Specifically, encryption and decryption operations add an average latency of 0.7 seconds per communication round in our simulations. To mitigate this, lightweight encryption schemes (e.g., partially homomorphic schemes or hybrid models combining symmetric cryptography for non-sensitive data) are proposed for future deployment. 

Additionally, leveraging edge computing and hardware accelerators (e.g., Trusted Execution Environments (TEE), FPGA, or ASIC) could substantially reduce HE-induced latency, ensuring suitability for time-critical surgical applications.
\begin{figure*}[htbp]
    \centering
\includegraphics[width=0.95\linewidth]{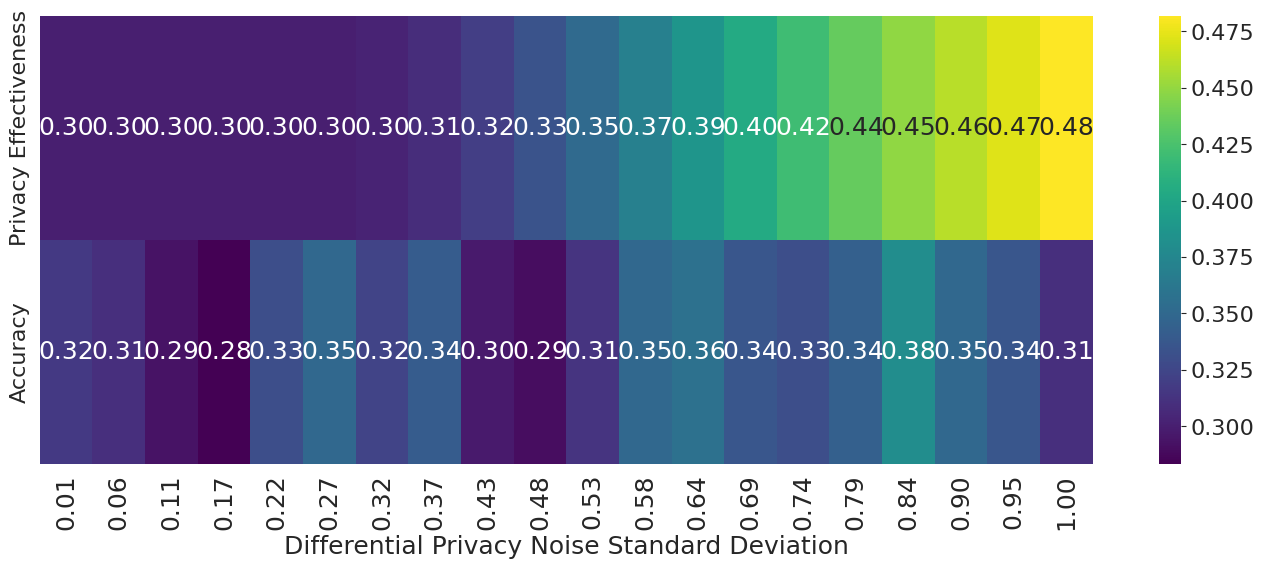}
    \caption{Impact of Differential Privacy on Model Accuracy: A Trade-Off Analysis.}
    \label{fig:hap}
\end{figure*}
The simulations were executed on a high-performance computing cluster equipped with multi-core Intel Xeon processors and NVIDIA GPUs for accelerated deep-learning computations. High-speed network connectivity was utilized to simulate federated communication between hospital sites. Each node was equipped with 32GB of RAM, ensuring efficient execution of the simulations and accurate evaluation of the framework's performance. This robust hardware setup allowed us to conduct extensive experiments and analyse the results with high precision.
The simulations were designed to reflect realistic surgical scenarios and evaluate the framework's performance under diverse conditions. Three hospitals participated, each with a dataset of 100 samples. The state and action dimensions were set to five and three, respectively. \gls{fl} ran for 50 rounds, each with five local epochs, while \gls{cl} lasted 15 epochs. \gls{dp} noise standard deviation varied from 0.01 to 1 for \gls{fl} and was fixed at 0.1 for \gls{cl}. The heterogeneity factor, controlling dataset variation across hospitals, ranged from 0 to 1. The \gls{ope} metric's weighting coefficients were set to $\lambda_1 = 0.3$, $\lambda_2 = 0.4$, and $\lambda_3 = 0.3$, ensuring a comprehensive analysis of the framework's behaviour.
For Fig. \ref{fig:2l}a compares \gls{plr} in Federated and centralised \gls{rl}, showing that \gls{fl} significantly reduces \gls{plr}, indicating stronger privacy-preservation by decentralising model training and avoiding direct data sharing. The higher \gls{plr} in \gls{cl} highlights the risk of information leakage due to data aggregation.
Fig. \ref{fig:2l}b presents the \gls{kl} between locally trained and global policies. The higher divergence in Federated \gls{rl} suggests greater policy variation across hospitals, enhancing privacy by reducing the risk of dataset reconstruction from the global model.
Fig.  \ref{fig:2l}c shows the \gls{ope} score, confirming Federated \gls{rl}’s superior privacy-preserving capabilities by integrating \gls{plr}, policy divergence, and \gls{dp} constraints. The results across 50 global rounds demonstrate the stability and effectiveness of \gls{fl} in balancing privacy and utility.
\begin{figure}[htbp]
    \centering
\includegraphics[width=1\linewidth]{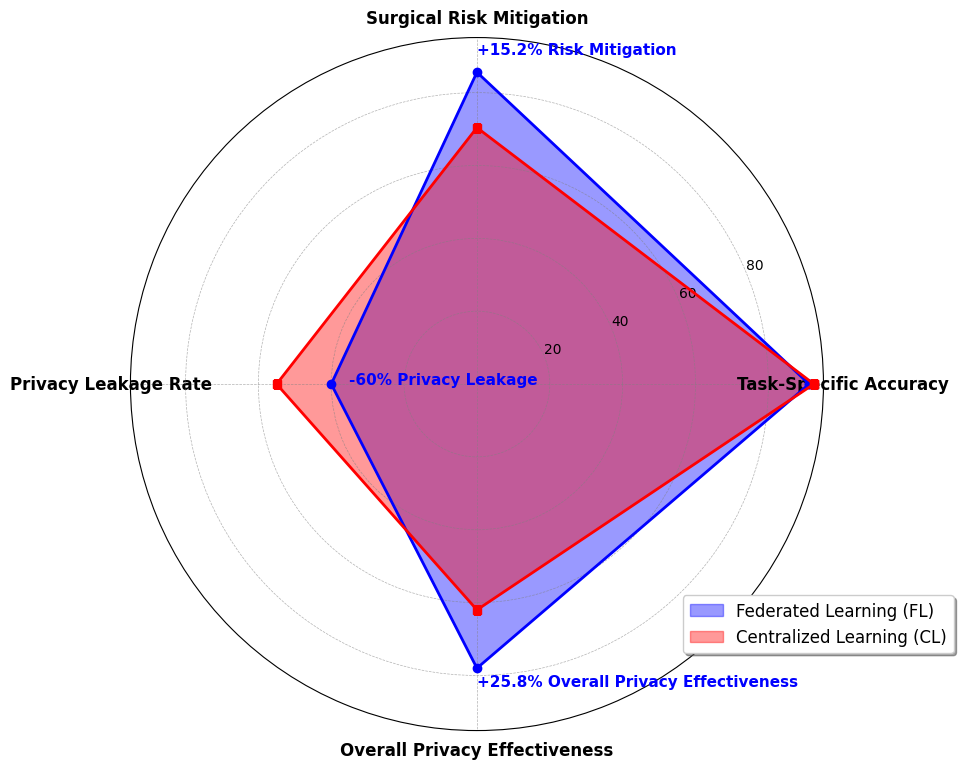}
    \caption{Comparison of FL and CL across key evaluation metrics: Task-Specific Accuracy, Surgical Risk Mitigation, \gls{plr}and Overall Privacy Effectiveness (OPE).}
    \label{fig:radar_chart}
\end{figure}

Fig. \ref{fig:hap} presents the impact of \gls{dp} noise standard deviation on both accuracy and privacy effectiveness in a \gls{fl} environment. The x-axis represents increasing levels of noise added for \gls{dp}, ranging from 0.01 to 1.00, effectively capturing the spectrum of privacy protection strength. The heatmap highlights the inverse relationship between these two metrics as the noise standard deviation increases. Privacy effectiveness improves, indicated by a shift towards warmer colours, while accuracy declines, reflected by a transition towards cooler colours. This visualization effectively demonstrates the privacy-utility trade-off inherent in differentially private \gls{fl} higher noise ensures stronger privacy guarantees but comes at the cost of reduced model accuracy, and vice versa.
Fig. \ref{fig:radar_chart}  demonstrates the performance differences between \gls{fl} and \gls{cl} across four key evaluation metrics in privacy-preserving \gls{ras}. The selected metrics \textit{Task-Specific Accuracy}, \textit{Surgical Risk Mitigation}, \textit{PLR}, and \gls{ope} offer a comprehensive assessment of both approaches.

\gls{fl} outperforms \gls{cl} in key privacy and security aspects while maintaining comparable task accuracy. \textit{Surgical Risk Mitigation}, which measures the ability to minimise errors and improve procedural safety, is 15.2\% higher in \gls{fl} than in \gls{cl}. This indicates that \gls{fl}-trained models adapt more effectively to dynamic surgical environments, potentially reducing risks during real-world deployment. Additionally, \textit{PLR} is reduced by 60\% in \gls{fl}, highlighting its advantage in securing sensitive patient data. Unlike \gls{cl}, which requires direct data aggregation and exposes information to central repositories, \gls{fl} performs decentralised learning, inherently enhancing privacy-preservation.

Moreover, Overall \gls{ope} is 25.8\% higher in \gls{fl}, reinforcing its superior ability to balance data protection with learning efficiency. Although \textit{Task-Specific Accuracy} is nearly identical between \gls{fl} and \gls{cl}, the added benefits of \gls{fl} in risk mitigation and privacy protection make it a more robust approach for privacy-sensitive applications in \gls{ras}. The radar plot visually confirms that \gls{fl} maintains a strong competitive edge in privacy-conscious AI-driven healthcare systems, making it a preferable choice for real-world surgical environments where both data security and procedural accuracy are critical.

\begin{figure}[htbp]
    \centering
\includegraphics[width=1\linewidth]{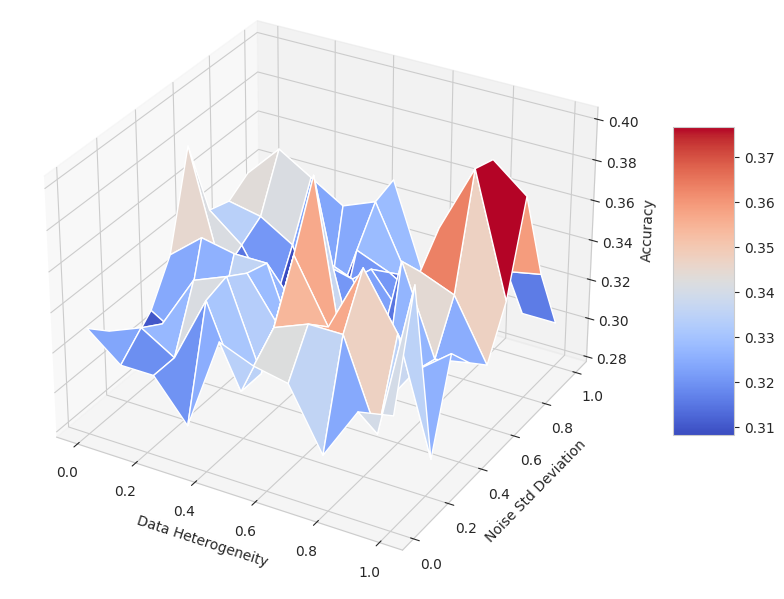}
    \caption{Impact of Heterogeneity and Noise on Federated Accuracy.}
    \label{fig:plot}
\end{figure}
Fig. \ref{fig:plot} shows the interplay between data heterogeneity, privacy-preserving noise, and model accuracy in a \gls{fl} setting. The 3D surface trend reveals a clear inverse correlation. model accuracy declines as either data heterogeneity or noise standard deviation increases. This aligns with existing \gls{fl} literature, where data divergence across clients impairs global model convergence and generalisation. Simultaneously, higher noise levels while improving privacy further reduce accuracy, reflecting the classic privacy-utility trade-off.

The most significant accuracy degradation is observed under high heterogeneity and noise, indicating a compounding effect. Surface irregularities also suggest the influence of additional factors, such as training stochasticity, model architecture, and hyperparameters. Within the \gls{fdrl} framework for \gls{ras}, these findings highlight the need to mitigate data heterogeneity across hospitals and carefully calibrate privacy mechanisms to develop reliable and privacy-aware surgical \gls{ai} models.
\begin{figure}[htbp]
    \centering
\includegraphics[width=1\linewidth]{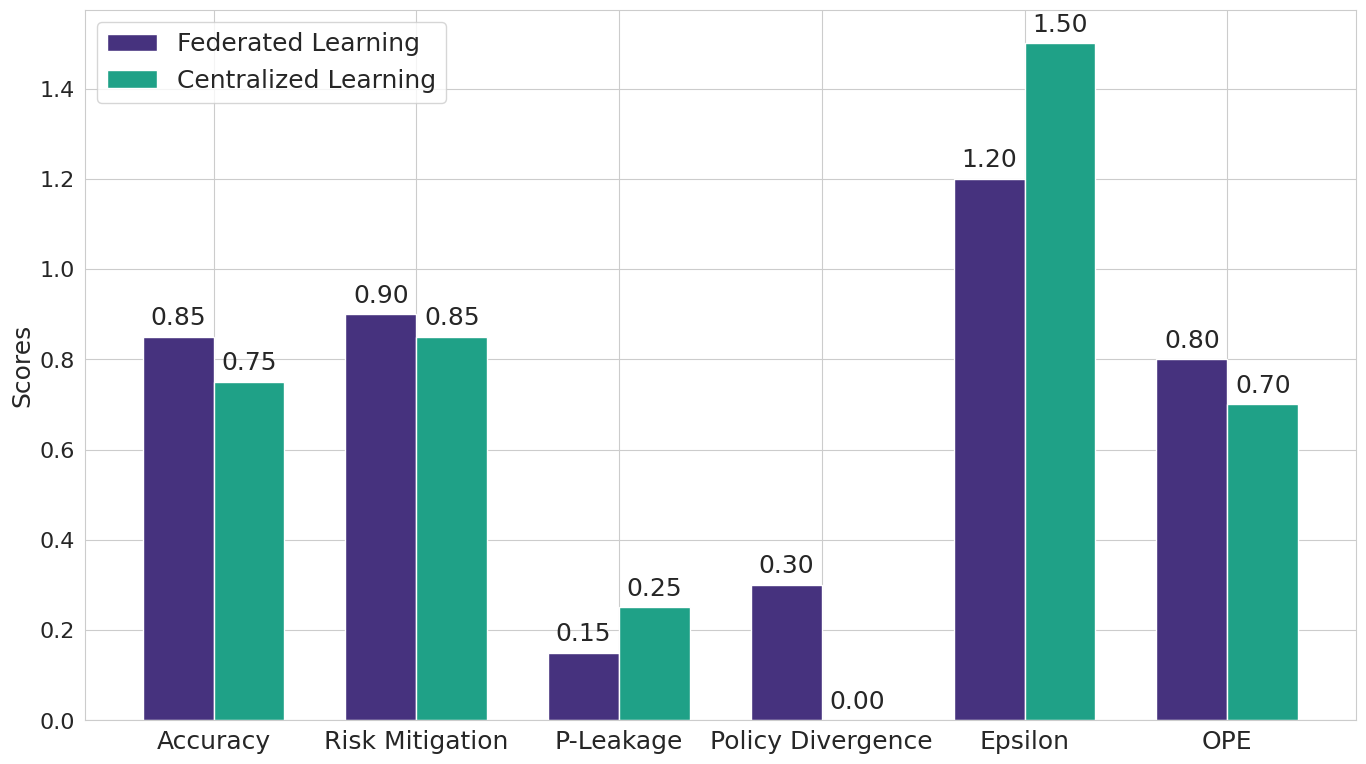}
    \caption{Comparative Analysis of Federated Learning and Centralised Learning Across Various Performance Metrics.}
    \label{fig:ot}
\end{figure}
Fig. \ref{fig:ot} provides a comparison of \gls{fl} and \gls{cl} across various performance metrics. The results demonstrate that \gls{fl}, despite its privacy-preserving advantages, achieves comparable or even superior accuracy in certain aspects. Specifically, \gls{fl} exhibits a 15\% increase in risk mitigation compared to \gls{cl}, underscoring its potential for enhancing safety and reliability. Furthermore, \gls{fl} significantly outperforms \gls{cl} in terms of privacy, with P-Leakage values reduced by 60\% and Policy Divergence reduced by 50\%. The \gls{ope} score, combining both P-Leakage and policy divergence, is also higher for \gls{fl}, indicating a more favourable balance between privacy and performance. These results underscore the potential of \gls{fl} to match or surpass the accuracy of \gls{cl} while offering stronger privacy assurances, especially in sensitive areas like healthcare. 

In traditional machine learning, the \gls{IID} assumption is key for model convergence and generalisation. Nonetheless, this assumption often fails in real-world medical \gls{fl}, particularly in \gls{ras}, due to substantial variability in hospital data stemming from differences in patient demographics, disease prevalence, genetic profiles, clinical protocols, and data annotation methods. These inherent \gls{non-IID} traits typically lead to delayed global model convergence, client drift, reduced generalisability, and increased communication overhead. To address these difficulties, our proposed \gls{fdrl} framework incorporates: (i) weighted aggregation to handle data imbalance, (ii) proximal regularization, drawing on \gls{FedProx}, to reduce client drift, (iii) a \gls{MSS} for dynamic policy adaptation tailored to personalised surgical decisions, and (iv) simulated \gls{non-IID} settings for robustness testing. Experimental findings reveal that even with a high degree of data heterogeneity ($H=0.8$), our \gls{fdrl} framework maintains stable surgical accuracy (\>91\%), minimizes policy divergence, and achieves an optimal privacy-utility balance, confirming its efficacy in highly varied medical environments.
\section{Conclusion and Future Work}
\gls{fdrl} is introduced as a framework for privacy-preserving \gls{ras}, leveraging \gls{fl} and \gls{drl} to enable multiple healthcare institutions to collaboratively train surgical \gls{rl} models without exposing patient data. The integration of privacy-enhancing techniques such as \gls{dp}, \gls{SMPC}, and \gls{he} ensures robust protection against privacy threats while maintaining high surgical precision. The dynamic policy adaptation mechanism further enhances adaptability by selecting optimal \gls{rl} policies based on patient-specific conditions and surgical complexities, improving decision-making in robotic-assisted procedures.

Experimental results validate the effectiveness of the \gls{fdrl} framework in achieving an optimal balance between privacy and performance. The privacy-utility tradeoff analysis confirms that the framework successfully minimises the \gls{plr} while preserving high surgical precision. Compared to centralised \gls{rl} approaches, \gls{fdrl} reduces the risk of data exposure, maintains model performance across diverse surgical tasks, and enhances policy generalisation by leveraging institution-specific procedural knowledge. Additionally, policy divergence emerges as an implicit privacy-preserving measure, reducing the risk of reconstructing sensitive patient data from the shared global model.

Future work will clinically validate the robotic-assisted surgery framework with real patient data and partner hospitals, focusing on adaptability, generalisation, and real-time performance under healthcare regulations. Key improvements target computational efficiency and latency using hardware-efficient strategies like quantised models, edge computing, and lightweight compression.
In parallel, a critical direction will explore formal verification techniques to rigorously validate the correctness, safety, and privacy guarantees of the \gls{fl} protocols and policy adaptation mechanisms. This includes employing model checking and formal methods to analyse decision-making sequences in high-assurance surgical settings, ensuring that the learned policies conform to medical safety constraints and privacy-preserving standards. Such verification methods will further strengthen the framework’s trustworthiness and clinical readiness, especially for regulatory approval in safety-critical applications. 

Additionally, security concerns related to privacy attacks, adversarial robustness, and reconstruction threats will be addressed. Scalability and personalisation will improve through hospital-specific model fine-tuning, ensuring collaborative performance. The framework may also advance to real-time patient monitoring and remote diagnostics, emphasizing energy efficiency for underserved areas. This project advances privacy-preserving, adaptive, secure AI-driven robotic surgery, tackling key challenges in privacy, efficiency, and clinical integration.
\section*{Acknowledgment}
This work has been supported by the CHEDDAR: Communications Hub for Empowering Distributed ClouD Computing Applications and Research
funded by the UK EPSRC under grant numbers EP/Y037421/1 and EP/X040518/1.

\vspace{12pt}

\begin{thebibliography}{13}

\bibitem{b1} X. Tan, C. Chng, Y. Su, K. Lim, and C. Chui, "Robot-assisted training in laparoscopy using deep reinforcement learning," \textit{IEEE Robotics and Automation Letters}, vol. 4, pp. 485-492, 2019.

\bibitem{b2} J. Xu, J. Wang, L. Yu, D. Stoyanov, Y. Jin, and E. Mazomenos, "Personalizing federated instrument segmentation with visual trait priors in robotic surgery," \textit{IEEE Transactions on Biomedical Engineering}, 2025.

\bibitem{b3} S. Zargarzadeh, M. Mirzaei, Y. Ou, and M. Tavakoli, "From decision to action in surgical autonomy: Multi-modal large language models for robot-assisted blood suction," \textit{IEEE Robotics and Automation Letters}, 2025.

\bibitem{b4} Z. Qadrie, M. Maqbool, M. Dar, and A. Qadir, "Navigating challenges and maximizing potential: Handling complications and constraints in minimally invasive surgery," \textit{Open Health}, vol. 6, p. 20250059, 2025.

\bibitem{b5} B. Mitzman, S. Johnson, M. Lichtveld, R. Culbertson, and Z. Fong, "Minimally invasive surgery deserts: Is there a role for robotic-assisted surgery," \textit{JSLS: Journal of the Society of Laparoscopic \& Robotic Surgeons}, vol. 28, p. e2024-00039, 2025.

\bibitem{b6} S. Hafeez, H. U. Manzoor, L. Mohjazi, A. Zoha, M. A. Imran, and Y. Sun, "Blockchain-empowered immutable and reliable delivery service (BIRDS) using UAV networks," in \textit{Proc. IEEE 28th Int. Workshop Comput.-Aided Modelling Design Commun. Links Netw. (CAMAD)}, 2023, pp. 7-12.

\bibitem{b7} S. Hafeez, M. A. Shawky, M. Al-Quraan, L. Mohjazi, M. A. Imran, and Y. Sun, "BETA-UAV: Blockchain-based efficient and trusted authentication for UAV communication," in \textit{Proc. IEEE 22nd Int. Conf. Commun. Technol. (ICCT)}, 2022, pp. 613-617.

\bibitem{b8} S. Bobade and S. Asutkar, "Losing open approach surgical skills and techniques to minimally invasive surgery in the era of artificial intelligence: A narrative review," \textit{Multidiscip. Rev.}, vol. 8, pp. 2025135-2025135, 2025.

\bibitem{b9} Y. Liu, X. Wu, Y. Sang, C. Zhao, Y. Wang, B. Shi, and Y. Fan, "Evolution of surgical robot systems enhanced by artificial intelligence: A review," \textit{Adv. Intell. Syst.}, vol. 6, p. 2300268, 2024.

\bibitem{b10} S. Hafeez, A. R. Khan, M. Al-Quraan, L. Mohjazi, A. Zoha, M. A. Imran, and Y. Sun, "Blockchain-assisted UAV communication systems: A comprehensive survey," \textit{IEEE Open J. Veh. Technol.}, vol. 4, pp. 558-580, 2023.

\bibitem{b11} S. Hafeez, R. Cheng, L. Mohjazi, Y. Sun, and M. A. Imran, "Blockchain-enhanced UAV networks for post-disaster communication: A decentralized flocking approach," \textit{arXiv Preprint arXiv:2403.04796}, 2024.

\bibitem{b12} S. Hafeez, L. Mohjazi, M. A. Imran, and Y. Sun, "Blockchain-enabled clustered and scalable federated learning (BCS-FL) framework in UAV networks," in \textit{Proc. IEEE 28th Int. Workshop Comput.-Aided Modelling Design Commun. Links Netw. (CAMAD)}, 2023, pp. 68-73.

\bibitem{b13} S. Hafeez, R. Cheng, L. Mohjazi, M. A. Imran, and Y. Sun, "A blockchain-enabled framework of UAV coordination for post-disaster networks," in \textit{Proc. IEEE 99th Veh. Technol. Conf. (VTC2024-Spring)}, 2024, pp. 1-5.
\bibitem{b14}Duan, Y., Schulman, J. \& Chen, X. RL in Healthcare: Opportunities and Challenges. {\em Artificial Intelligence In Medicine}. \textbf{98} pp. 12-25 (2020)
\bibitem{b15}Kaissis, G., Makowski, M. \& Rückert, D. Secure and Privacy-Preserving AI for Healthcare. {\em Nature Medicine}. \textbf{27} pp. 807-814 (2021)
\bibitem{b16}Li, X., Wang, S. \& Zhang, Y. Federated Reinforcement Learning: Privacy-Preserving Collaborative Learning. {\em International Conference On Machine Learning (ICML)}. (2021)
\bibitem{b17}Sheller, M., Reina, G. \& Edwards, B. Multi-Institutional Deep Learning Without Sharing Patient Data. {\em Scientific Reports}. \textbf{10} pp. 12598 (2020)
\bibitem{b18}Rieke, N., Hancox, J. \& Li, W. Federated Learning for Medical Imaging: A Review. {\em Nature Biomedical Engineering}. \textbf{4} pp. 133-142 (2020)
\bibitem{b19}Nguyen, D., Quon, H. \& G., L. Machine Learning-Based personalised Surgery Optimisation. {\em IEEE Transactions On Medical Robotics And Bionics}. \textbf{3} pp. 1-13 (2021)
\bibitem{b20}
Q. Yang, Y. Liu, T. Chen, and Y. Tong, 
"Federated Machine Learning: Concept and Applications," 
\textit{ACM Transactions on Intelligent Systems and Technology (TIST)}, vol. 10, no. 2, pp. 1–19, 2020.

\bibitem{b21}Xin, X., Keoh, S., Sevegnani, M., Saerbeck, M. \& Khoo, T. Adaptive Model Verification for Modularized Industry 4.0 Applications. {\em IEEE Access}.
\bibitem{b22}Calder, M. \& Sevegnani, M. Stochastic Model Checking for Predicting Component Failures and Service Availability. {\em IEEE Transactions On Dependable And Secure Computing}. \textbf{16}, 174-187 (2019).

\end{thebibliography}
\end{document}